\documentclass[letterpaper, 10pt, conference]{ieeeconf}

\IEEEoverridecommandlockouts
\usepackage[utf8]{inputenc}
\usepackage{graphicx}
\usepackage{tikz}
\usepackage{xcolor}
\usetikzlibrary{positioning, arrows.meta, fit, backgrounds, calc}
\usepackage{amsmath,amssymb}
\usepackage{booktabs}
\let\labelindent\relax
\usepackage{enumitem}
\usepackage{hyperref}

\title{\LARGE \bf
TactileEval: A Step Towards Automated Fine-Grained Evaluation and Editing
of Tactile Graphics
}

\author{Adnan Khan$^{1,*}$,
Abbas Akkasi$^{1}$ and
Majid Komeili$^{1}$%
\thanks{$^{1}$School of Computer Science, Carleton University, Ottawa, Canada}%
\thanks{$^{*}$Corresponding author: Adnan Khan. E-mail: adnankhan5@cmail.carleton.ca}%
}

\begin{document}

\maketitle
\thispagestyle{empty}
\pagestyle{empty}

\begin{abstract}
Tactile graphics require careful expert validation before reaching blind and
visually impaired (BVI) learners, yet existing datasets provide only coarse
holistic quality ratings that offer no actionable repair signal.
We present \textsc{TactileEval}, a three-stage pipeline that takes a first
step toward automating this process.
Drawing on expert free-text comments from the TactileNet dataset, we establish a five-category quality taxonomy; encompassing view angle, part completeness, background clutter, texture separation, and line quality—aligned with BANA standards. We subsequently gathered 14,095 structured annotations via Amazon Mechanical Turk, spanning 66 object classes organized into six distinct families.
A reproducible ViT-L/14 feature probe trained on this data achieves 85.70\%
overall test accuracy across 30 different tasks, with consistent difficulty
ordering suggesting the taxonomy suggesting the taxonomy captures meaningful perceptual structure.
Building on these evaluations, we present a ViT-guided automated editing
pipeline that routes classifier scores through family-specific prompt templates
to produce targeted corrections via \texttt{gpt-image-1} image editing.
Code, data, and models are available at \url{https://TactileEval.github.io/}.
\end{abstract}

\section{Introduction}
\label{sec:intro}

Tactile graphics are embossed representations of visual content that allow BVI
individuals to perceive images through touch.
Producing high-quality tactile graphics is labor-intensive: experts must inspect
line quality, texture encoding, posture, background clutter, and viewpoint
consistency before embossed artifacts reach BVI learners
\cite{mukhiddinov2021systematic, vcervenka2013tactile}.
TactileNet~\cite{khan2025tactilenet} represents an important advance by curating
natural-photo/tactile-drawing pairs with expert quality ratings; however, those
ratings are coarse: four holistic levels (Accept as Is, Minor Edit, Major Edit, and Reject). These do not specify \emph{which}
perceptual attributes are flawed nor \emph{how} they should be corrected.

This paper presents \textsc{TactileEval}, a pipeline addressing this gap through
three complementary contributions:
\begin{itemize}[leftmargin=*]
    \item \textbf{A fine-grained quality dataset} spanning 30 distinct task families; derived from the intersection of six object families and five BANA-aligned quality dimensions. The corpus includes 14,095 structured annotations across 66 TactileNet object classes, collected via a rigorous AMT protocol (Sections~\ref{sec:taxonomy}--\ref{sec:dataset}).
    \item \textbf{A reproducible ViT-based quality evaluator} that achieves
          85.70\% overall test accuracy across 30 different tasks, providing a
          structured alternative to expert-only annotation
          (Section~\ref{sec:vit}).
    \item \textbf{A ViT-guided editing pipeline} that translates automated
          quality scores into targeted image edits via family-specific
          prompt templates, taking a concrete step toward end-to-end tactile
          repair (Section~\ref{sec:editing}).
\end{itemize}


\begin{figure}[t]
\centering
\begin{tikzpicture}[
    node distance = 0.38cm,
    box/.style  = {draw, rounded corners=3pt,
                   minimum width=4.6cm, minimum height=0.52cm,
                   font=\footnotesize, align=center, fill=white},
    hdr/.style  = {draw, rounded corners=3pt,
                   minimum width=4.6cm, minimum height=0.52cm,
                   font=\footnotesize\bfseries, align=center,
                   fill=blue!12},
    arr/.style  = {-{Latex[length=3.5pt,width=3.5pt]}, thick},
    grp/.style  = {draw=gray!50, rounded corners=4pt,
                   fill=gray!7, inner sep=5pt}
]

\node[hdr]                    (A) {Expert Comments};
\node[box, below=of A]        (B) {Thematic Analysis \& Taxonomy (5 dims)};
\node[box, below=of B]        (C) {AMT HIT Design (per family $\times$ dimension)};
\node[box, below=of C]        (D) {Qualification + Gold-Sample QC};
\node[box, below=of D]        (E) {7-Worker Vote Aggregation};
\node[box, below=of E]        (F) {Consensus Filter (majority voting)};
\node[hdr, below=of F]        (G) {14,095 Binary Records (11,348 / 1,341 / 1,406)};

\draw[arr] (A) -- (B);
\draw[arr] (B) -- (C);
\draw[arr] (C) -- (D);
\draw[arr] (D) -- (E);
\draw[arr] (E) -- (F);
\draw[arr] (F) -- (G);

\begin{scope}[on background layer]
  \node[grp, fit=(A)(B)(C)(D)(E)(F)(G)] {};
\end{scope}

\end{tikzpicture}
\caption{Dataset construction pipeline.
Expert free-text comments from TactileNet seed a five-category quality
taxonomy, which drives AMT HIT design across all six object families.
Qualified workers complete HITs with gold-sample quality control; votes are
aggregated and filtered for consensus, yielding 14,095 binary records.}
\label{fig:pipeline}
\end{figure}
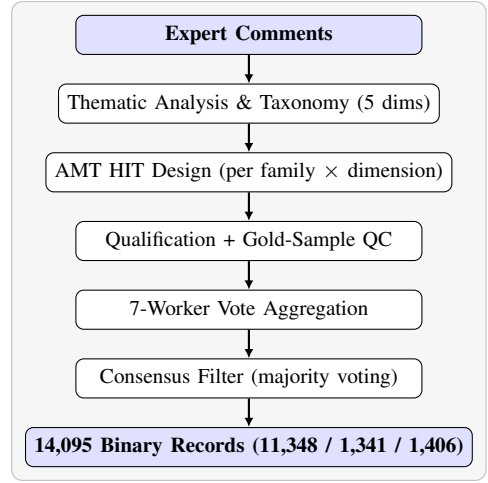
Fig.~\ref{fig:pipeline} illustrates the dataset construction stage.
The pipeline deliberately decomposes expert tactile knowledge into structured,
option-level judgments expressible in plain language; removing the reliance on
domain specialists for each new annotation cycle and enabling scalable quality
assessment across the full TactileNet catalog.

\section{Background and Related Work}
\label{sec:background}

\subsection{TactileNet Dataset}
TactileNet~\cite{khan2025tactilenet} is the first large-scale dataset for
generating embossing-ready tactile graphics, pairing natural photographs across
66 object classes with tactile line drawings at scale via Stable Diffusion
models fine-tuned with LoRA and DreamBooth.
Expert evaluation reported 92.86\% adherence to tactile accessibility standards,
with an SSIM of 0.538 between generated and expert-designed images.
While TactileNet demonstrates that generation is tractable, its expert quality
ratings remain holistic and do not identify which attributes fail nor prescribe
corrections/edits; the gap \textsc{TactileEval} addresses.

\subsection{Tactile Graphics for BVI Learners}
Tactile graphics are essential for conveying spatial and diagrammatic information
to BVI individuals~\cite{edman1992tactile}, relying on raised lines, textures,
and shapes to communicate visual structure through fingertip exploration.
Guidelines from BANA~\cite{bana2010guidelines} and RNIB~\cite{rnib2010guidelines}
codify best practices covering viewpoint simplification, texture separation, and
minimum stroke weight; criteria that directly inform our five quality dimensions.
Recent work has begun automating parts of this pipeline via rule-based vector
simplification~\cite{magne2025single} and neural map-to-tactile
translation~\cite{choubineh2025text}, but quality verification and correction
remain unsolved.

\subsection{Crowdsourced Quality Assessment}
Crowdsourced annotation via Amazon Mechanical Turk has been widely adopted for
perceptual image quality benchmarks~\cite{ponomarenko2015tid2013,lin2019kadid}.
Authors in~\cite{snow2008cheap} established that crowd labels can match
expert-level quality when majority-vote aggregation and worker filtering are
applied; principles we adopt directly.
Tactile graphics introduce additional requirements: workers must compare a
natural photograph with a line-art embossed drawing, demanding targeted training
and qualification gating beyond standard image quality tasks.

\subsection{Vision-Language Models for Visual Assessment}
Contrastive vision-language models such as CLIP~\cite{radford2021clip} build on
contrastive self-supervised learning~\cite{khan2022contrastive} to encode
features transferable to downstream classification via lightweight probing.
MLLMs including GPT-4o~\cite{openai2023gpt4} and LLaVA~\cite{liu2023llava}
achieve strong performance on visual question answering through instruction
tuning. No prior work applies either paradigm to tactile or
accessibility-specific quality assessment.

\section{Quality Taxonomy}
\label{sec:taxonomy}

\subsection{Mining TactileNet Expert Comments}
TactileNet's annotation form included an optional free-text field describing perceptual problems in each tactile drawing. We performed a qualitative thematic analysis of approximately 80 non-empty expert comments from ``non-accept'' image pairs to surface recurring tactile failures. This analysis formed the empirical basis for our structured taxonomy.

\subsection{Five BANA-Aligned Quality Dimensions}
\label{sec:dimensions}

To convert expert feedback into a scalable benchmarking framework, we established five quality dimensions grounded in BANA standards: \textbf{View Match} (V), \textbf{Required Parts} (P), \textbf{Identity \& Background} (B), \textbf{Texture Separation} (T), and \textbf{Line Quality} (L). 

We mapped these dimensions across six object families (Animals, Food/Nature, Vehicles, etc.), resulting in 30 distinct task families (F\textit{n}Q\textit{X}). To ensure high-quality crowdsourced labels, we adapted the instructions for each family to use domain-specific terminology. For example, while the \textbf{Required Parts (P)} dimension is conceptually identical across the dataset, the prompt for Animal families (F1) focuses on anatomical features like limbs and tails, whereas for Simple Objects (F2), it targets functional components like stems, handles, or support bases.

Each task is presented as a multi-select technical probe rather than a binary check, requiring workers to identify specific failure modes:

\begin{itemize}[leftmargin=*]
    \item \textbf{View Match (QV):} Compares tactile orientation (e.g., Front, Side, Top) against the reference photo to prevent perspective-driven confusion.
    \item \textbf{Required Parts (QP):} Identifies missing, hallucinated, or incorrectly placed components relative to the object's canonical structure.
    \item \textbf{Identity \& Background (QB):} Flags background clutter, extraneous artifacts, or failures in preserving the object's semantic class (e.g., mismatched species or vehicle type).
    \item \textbf{Texture Separation (QT):} Evaluates tactile differentiation between adjacent regions, flagging inconsistent patterns, boundary crossings, or excessive density.
    \item \textbf{Line Quality (QL):} Assesses the physical integrity of the drawing's outlines, specifically identifying broken, fuzzy, or colliding bold strokes.
\end{itemize}

Table~\ref{tab:taxonomy_map} demonstrates how these generic dimensions were operationalized using family-specific prompts.

\begin{table}[h]
\caption{Operationalization of quality dimensions across different families}
\label{tab:taxonomy_map}
\renewcommand{\arraystretch}{1.1}
\begin{center}
\begin{tabular}{lp{0.35\columnwidth}p{0.35\columnwidth}}
\toprule
\textbf{Dimension} & \textbf{Animal (F1) Context} & \textbf{Food/Objects (F2) Context} \\
\midrule
\textbf{Parts (P)} & Missing limbs, tails, ears & Missing stems, leaves, handles \\
\textbf{Texture (T)} & Fur vs. wing differentiation & Skin vs. seed differentiation \\
\textbf{Identity (B)} & Species/breed accuracy & Type (e.g., fruit vs. tool) accuracy \\
\textbf{Lines (L)} & Solid, traceable contours & Solid, traceable contours \\
\bottomrule
\end{tabular}
\end{center}
\end{table}
Table~\ref{tab:options} lists all checkbox options per quality dimension.
Each dimension contains between 3 and 7 options; F\textit{n}QT is the most
granular, reflecting the subjective nature of texture judgments.

\begin{table}[h]
\caption{Checkbox options per quality dimension and their focus}
\label{tab:options}
\renewcommand{\arraystretch}{1.1}
\begin{center}
\begin{tabular}{p{0.10\columnwidth}p{0.78\columnwidth}}
\toprule
\textbf{Dim.} & \textbf{What workers assess} \\
\midrule
QB & Species/object identity, pose or configuration match, background cleanliness \\
\addlinespace
QL & Broken or discontinuous outlines, overly bold strokes, blurry/low-contrast lines, or no issues \\
\addlinespace
QP & Missing anatomical parts, hallucinated structures, incorrect placement, or all correct \\
\addlinespace
QT & Absent texture, intra-element inconsistency, near/far confusion, region bleed, density, similarity to adjacent regions, or no issues \\
\addlinespace
QV & View-angle match; if mismatch: frontal, side, top, or perspective; F2 adds an option for geometrically ambiguous viewpoints \\
\bottomrule
\end{tabular}
\end{center}
\end{table}
\section{Dataset Construction}
\label{sec:dataset}

\subsection{AMT Data Collection Protocol}
\label{sec:amt}

Each Human Intelligence Task (HIT) presented workers with a side-by-side view
of a natural image and its corresponding tactile drawing.
Workers selected all applicable failure options from a dimension-specific
checkbox list, or indicated no issue if the tactile was satisfactory.
Annotation quality was enforced through a two-stage vetting process.

\paragraph{Training examples.}
Each task's instruction page presented three annotated reference pairs
representing the full spectrum: a clean tactile, a single clear defect, and a
pair with multiple co-occurring issues.

\paragraph{Gold-sample quality control.}
Each main HIT embedded at least two gold-sample questions with manually verified
labels.
Assignments failing the gold threshold were rejected and re-posted to maintain
annotation quality.

\paragraph{Qualification test.}
A custom qualification test per task family required workers to meet a 2/3
accuracy threshold; only one attempt was permitted per 24-hour window.
Task visibility was set to \emph{Private}: only qualified workers could accept
assignments.

\paragraph{Label aggregation.}
For each image pair and checkbox option, we compute the vote fraction across 7
workers.
An option is labeled \textsc{True} when the vote fraction meets
or exceeds a per-task confidence threshold: $\geq 0.6$ for F\textit{n}QV,
F\textit{n}QP, F\textit{n}QB, and F\textit{n}QL---the natural strict-majority
threshold for 7-worker HITs ($\lceil7/2\rceil = 4$ votes $\approx 0.571$,
rounded to 0.6)---and $> 0.4$ for F\textit{n}QT, where texture judgments
exhibited higher inter-worker variability and the stricter majority threshold
suppressed valid positive annotations.

\subsection{Full-Scale Dataset}
\label{sec:fulldata}

We organise the 66 TactileNet object classes into six broad families:
Animals \& Creatures (F1), Food \& Nature (F2), Furniture \&
Structures (F3), , Tools \& Instruments (F4), Vehicles \& Flight Systems (F5), and Wearables \& Accessories (F6).
The AMT collection was deployed across all six families and all five quality
dimensions, yielding 30 distinct task families.

Raw AMT results were reprocessed across all families with normalized option
schemas and a two-stage consensus filter: approved ballots are kept; additionally,
votes showing agreement among at least five workers on an identical label vector
are promoted, while under-supported or tied options are dropped.
The resulting dataset contains only consensus-backed binary decisions along with
exact vote counts per pair and option.

The final dataset comprises \textbf{14,095} option-level binary records split
into \textbf{11,348} training / \textbf{1,341} validation / \textbf{1,406} test
examples.
Each record contains the image pair identifiers, task family, option identifier
and description, majority label, vote fraction, vote counts, and provenance
metadata.
Because each image pair is evaluated across multiple options (one binary record
per option), the dataset captures co-occurring defects within a single
structured representation.

The F1QV prompt for Animals \& Creatures illustrates the plain-language design:

\begin{center}
\fcolorbox{gray!60}{gray!8}{%
  \parbox{0.85\columnwidth}{%
    \small\raggedright
    \textbf{F1QV prompt:} ``Does the animal show the same flat view as the
    photo (same head/body angle)?''\\[4pt]
    \textbf{Options:} \texttt{angle\_match} · \texttt{view\_frontal} ·
    \texttt{view\_side} · \texttt{top\_view} · \texttt{view\_perspective}
  }%
}
\end{center}

\section{ViT-Based Quality Evaluation}
\label{sec:vit}

\subsection{Feature Probe Architecture}

To provide a reproducible, cost-transparent evaluator, we train a ViT-based
feature probe that deliberately separates representation from decision-making.

\paragraph{Feature extraction.}
For each record, we extract image embeddings from both the natural and tactile
images using a frozen CLIP ViT-L/14 model~\cite{radford2021clip} initialized
from the \texttt{laion2b\_s34b\_b88k} pretrained checkpoint \cite{10.5555/3600270.3602103}.
A text embedding is extracted from the option prompt
\textit{``Task \{family\} option \{option\_id\}: \{description\}''}.
All embeddings are $\ell_2$-normalized; each image and text embedding is
768-dimensional, yielding a concatenated feature vector of
$4 \times 768 = 3{,}072$ dimensions.
The final vector concatenates: the natural image embedding, the tactile
image embedding, their element-wise difference (capturing cross-modal
discrepancy), and the text option embedding.

\begin{figure}[t]
    \centering
    \includegraphics[width=\columnwidth, height=8cm,
        keepaspectratio]{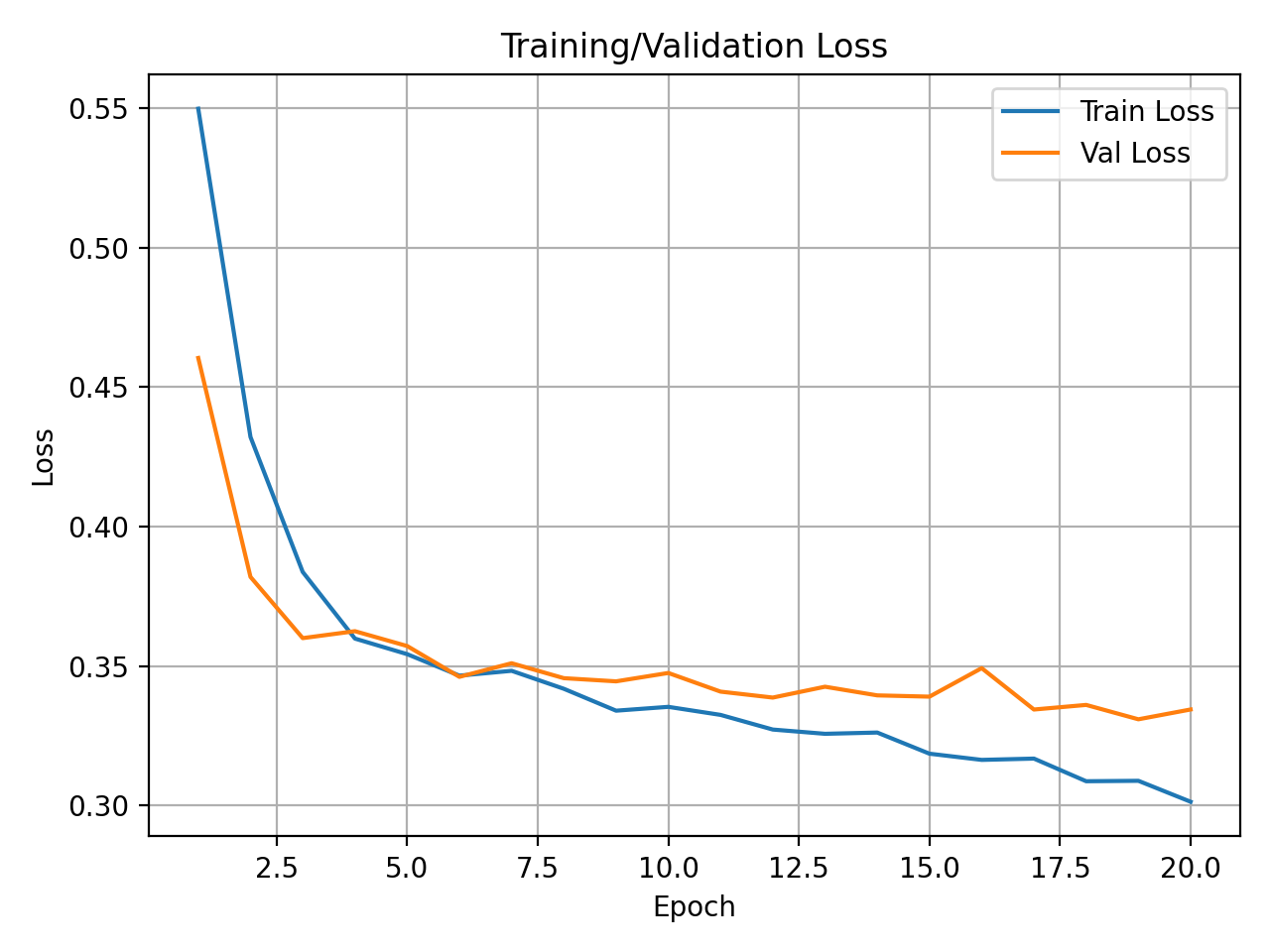}
    \caption{Training and validation loss over 20 epochs. Both curves converge
    smoothly with no divergence, indicating stable generalisation throughout
    training.}
    \label{fig:loss_curve}
\end{figure}

\paragraph{Classifier and training.}
Each quality option is treated as an independent binary classification problem
(issue present vs.\ absent), with the sigmoid output serving directly as an
issue probability. A softmax multi-class formulation over all options within a
task family was also explored but yielded sub-optimal results, likely because
options are not mutually exclusive and co-occurring defects are common.
A two-layer MLP (Linear(3072, 512)--ReLU--Linear(512, 1)) is trained per option
using binary cross-entropy loss and AdamW (lr $1{\times}10^{-3}$) with batch
size 128 over 20 epochs, selecting the best checkpoint by peak validation
accuracy. The CLIP ViT-L/14 backbone is fully frozen throughout;
only the MLP weights are updated.
All experiments run on a single NVIDIA RTX~4090 (24\,GB VRAM).
\paragraph{Training dynamics.}
Fig.~\ref{fig:loss_curve} shows the training and validation loss curves.
Both losses drop sharply in the first three epochs and converge smoothly
thereafter, with no sign of overfitting: validation loss tracks and slightly
leads training loss in early epochs before the two stabilise in parallel,
suggesting the model generalises well to held-out pairs.

\subsection{Evaluation Results}
\label{sec:results}

Table~\ref{tab:test_results} reports per-family test accuracy across all six
object families.
The probe achieves \textbf{85.70\%} overall test accuracy on 1,406 test records.
Family-level accuracy ranges from 79.51\% (F6, Food \& Nature) to 89.31\%
(F2, Vehicles \& Flight Systems).

\begin{table}[h]
\caption{ViT feature probe test accuracy across all six families}
\label{tab:test_results}
\renewcommand{\arraystretch}{1.1}
\begin{center}
\begin{tabular}{lll}
\toprule
\textbf{Family} & \textbf{Object Category} & \textbf{Acc.\ (\%)} \\
\midrule
F1 & Animals \& Creatures      & 84.72 \\
F2 & Vehicles \& Flight        & 89.31 \\
F3 & Furniture \& Structures   & 87.25 \\
F4 & Wearables \& Accessories  & 87.86 \\
F5 & Tools \& Instruments      & 80.00 \\
F6 & Food \& Nature            & 79.51 \\
\midrule
\multicolumn{2}{l}{\textbf{Overall}}   & \textbf{85.70} \\
\bottomrule
\end{tabular}
\end{center}
\end{table}

\begin{figure}[htbp]
    \centering
    \includegraphics[width=\columnwidth]{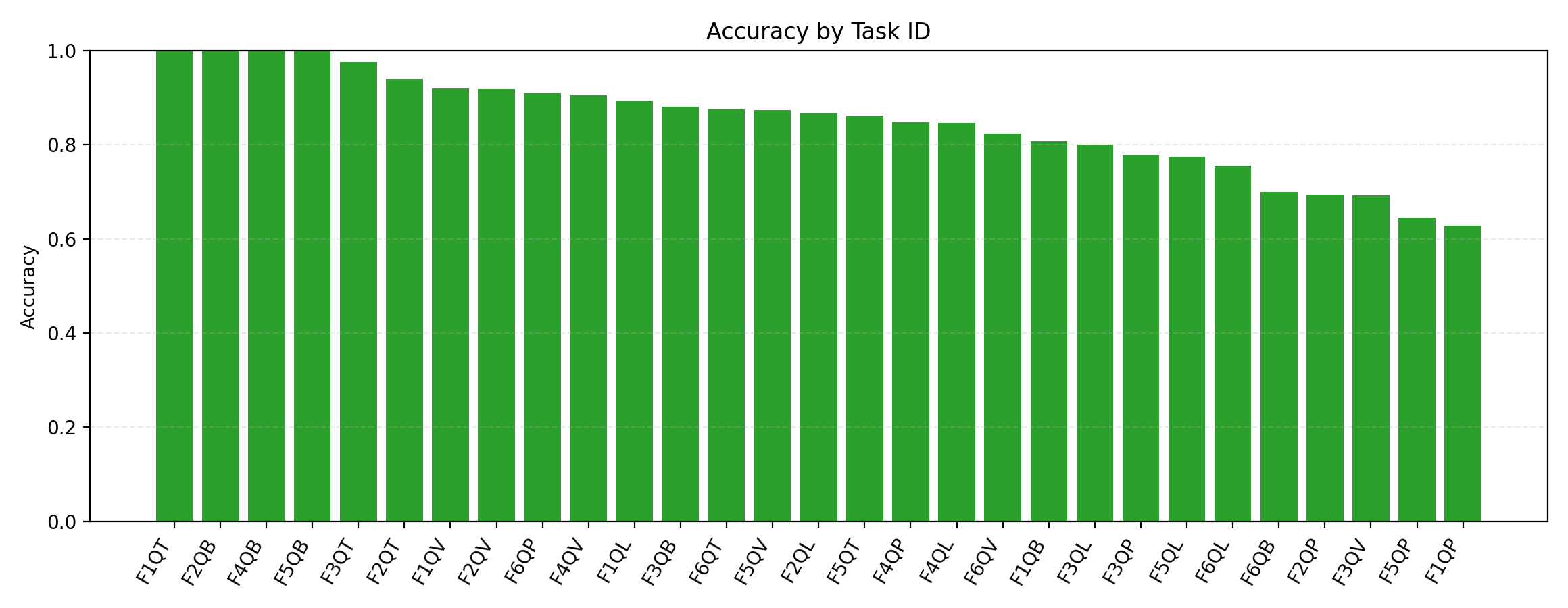}
    \caption{ViT feature probe test-set accuracy across all 30 tasks (sorted descending). Background checks (F2QB, F4QB, F5QB) reach 1.0, while anatomical and texture checks (F1QP, F5QP) prove the most challenging.}
    \label{fig:vit_results_task}
\end{figure}

\begin{figure}[htbp]
    \centering
    \includegraphics[width=\columnwidth]{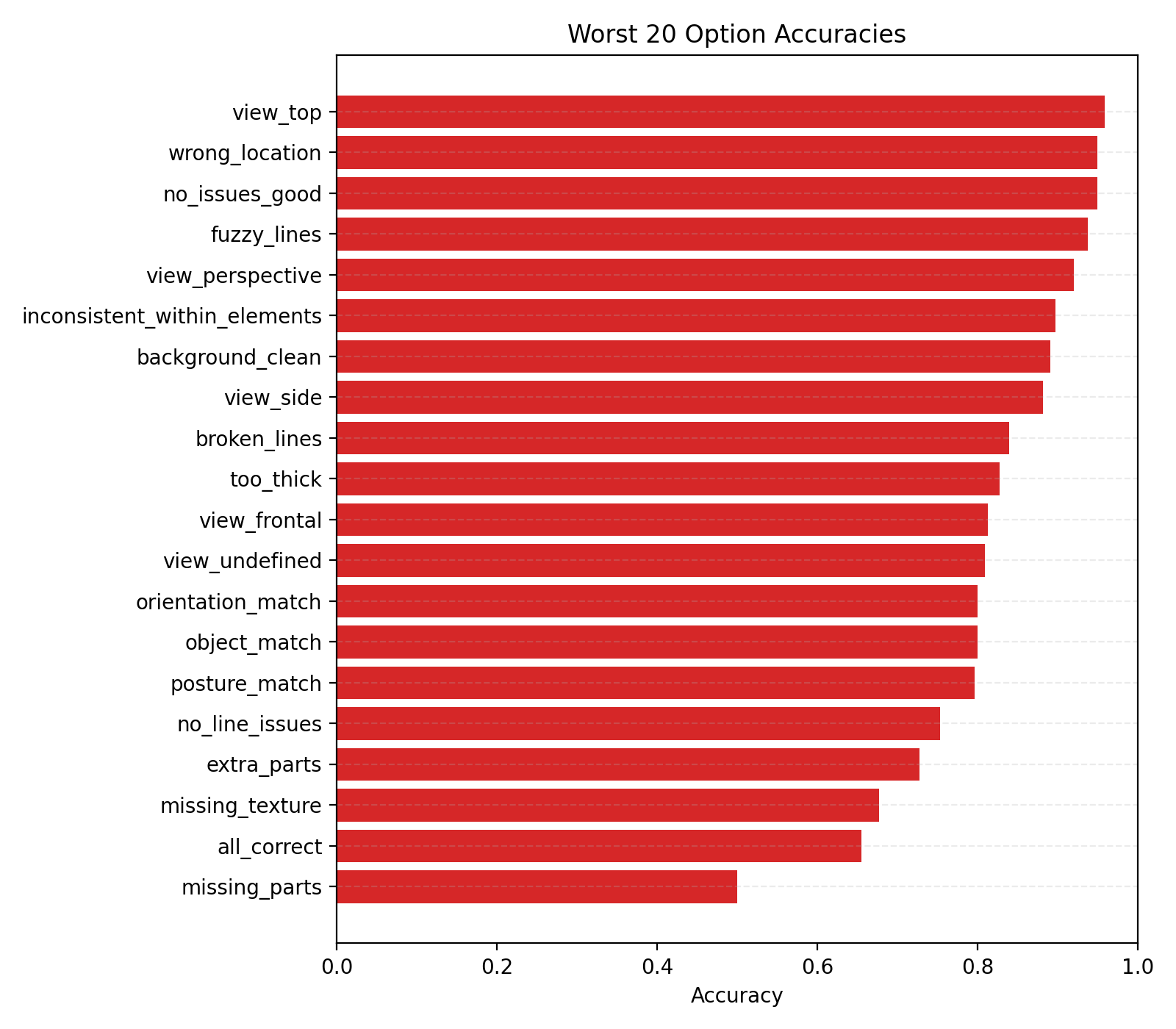}
    \caption{~Bottom-20 option accuracies (per-option accuracy is the fraction of correctly predicted samples for that option independently; options are not macro-averaged).}
    \label{fig:vit_results_option}
\end{figure}

Fig.~\ref{fig:vit_results_task} visualizes accuracy across all 30 task families. Tasks such as F2QB, F4QB, and F5QB (background/extra-content checks) reach perfect accuracy, consistent with background cleanliness being a visually unambiguous signal. Challenging tasks concentrate in F1QP and F5QP (required parts), where minimalistic tactile drawings can legitimately omit detail, making the part-presence judgment inherently subjective.

At the option level (Fig.~\ref{fig:vit_results_option}), the hardest cases include \texttt{missing\_parts} (0.50), and \texttt{missing\_texture} (0.68); requiring localized, fine-grained reasoning about fill density or anatomical completeness that global ViT embeddings struggle to resolve. Geometric and identity checks such as \texttt{angle\_match}, \texttt{bleeds\_boundaries}, and \texttt{configuration\_match} reach 1.0, indicating that the difficulty ordering reflects genuine task structure rather than annotation artefacts.
\section{ViT-Guided Editing Pipeline}
\label{sec:editing}

While the ViT evaluator identifies \emph{what} may be wrong with a tactile
graphic, correcting identified defects requires translating classifier scores into targeted image
edits.
We present a ViT-guided editing pipeline as a concrete step in this direction,
with the explicit acknowledgment that a comprehensive tactile correction system
remains an open challenge.
The pipeline currently addresses one diagnosed issue per invocation; a future
``TactileExpert Editor'' that jointly resolves all defects in a single pass is a
natural extension left for future work.

\subsection{Pipeline Design}
\label{sec:edit_design}


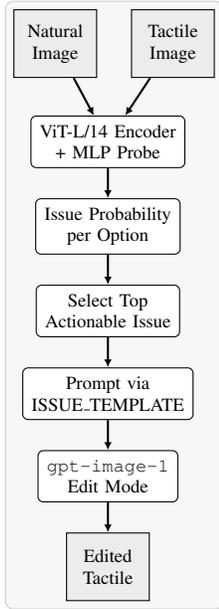
\begin{figure}[t]
\centering
\begin{tikzpicture}[
    node distance = 0.35cm and 0.45cm,
    img/.style  = {draw, minimum width=1.1cm, minimum height=0.9cm,
                   fill=gray!15, font=\scriptsize, align=center},
    box/.style  = {draw, rounded corners=2pt, minimum width=1.5cm,
                   minimum height=0.5cm, font=\scriptsize, align=center,
                   fill=white},
    arr/.style  = {-{Latex[length=3pt,width=3pt]}, thick},
    grp/.style  = {draw=gray!40, rounded corners=3pt,
                   fill=gray!6, inner sep=3pt}
]

\node[img] (nat) {Natural\\Image};
\node[img, right=of nat] (tac) {Tactile\\Image};

\node[box, below=0.5cm of nat, xshift=0.7cm] (enc) {ViT-L/14 Encoder\\+ MLP Probe};

\draw[arr] (nat) -- (enc);
\draw[arr] (tac) -- (enc);

\node[box, below=0.4cm of enc] (ip) {Issue Probability\\per Option};
\draw[arr] (enc) -- (ip);

\node[box, below=0.4cm of ip] (sel) {Select Top\\Actionable Issue};
\draw[arr] (ip) -- (sel);

\node[box, below=0.4cm of sel] (pt) {Prompt via\\ISSUE\_TEMPLATE};
\draw[arr] (sel) -- (pt);

\node[box, below=0.4cm of pt] (edit) {\texttt{gpt-image-1}\\Edit Mode};
\draw[arr] (pt) -- (edit);

\node[img, below=0.4cm of edit] (out) {Edited\\Tactile};
\draw[arr] (edit) -- (out);

\begin{scope}[on background layer]
  \node[grp, fit=(nat)(tac)(enc)(ip)(sel)(pt)(edit)(out)] {};
\end{scope}

\end{tikzpicture}
\caption{ViT-guided editing pipeline.
The natural and tactile images are encoded by the frozen ViT-L/14 backbone;
the MLP probe assigns issue probabilities per option.
The top actionable issue maps to a family-specific repair instruction that is
submitted to \texttt{gpt-image-1} in edit mode, constraining corrections to the
existing drawing.}
\label{fig:pipeline_editing}
\end{figure}

For a given tactile image and target task family, the pipeline proceeds as
follows (Fig.~\ref{fig:pipeline_editing}).

\paragraph{Issue scoring.}
The natural image, tactile image, their element-wise embedding difference, and
each option's text prompt are encoded by the frozen ViT-L/14 backbone and the
trained MLP head.
The resulting \emph{issue probability} inverts the raw sigmoid output for
negative-polarity pass signals (e.g., \texttt{no\_line\_issues}), so that all
scores express the probability of a defect.
The highest-scoring actionable option is selected as the repair target.

\paragraph{Prompt construction.}
The selected issue code is mapped to a natural-language repair instruction via
\texttt{ISSUE\_TEMPLATES}.
This is assembled into a structured prompt with a family-level header and footer
from \texttt{FAMILY\_PROMPT\_FRAMES}, which enforce tactile formatting
constraints (clean silhouettes, stroke continuity, background cleanliness) as
guardrails independent of the specific defect.

\paragraph{Image editing.}
The original tactile is padded to a square canvas and submitted as the base
image to \texttt{gpt-image-1} in edit mode, so edits are applied to the
existing drawing rather than generating from scratch.
The pipeline saves the prompt, edited image, and structured metadata for human
review.

\subsection{Zero-Shot Baseline}
\label{sec:zeroshot}

\begin{figure}[t]
    \centering
    \begin{minipage}[t]{0.48\columnwidth}
        \centering
        \includegraphics[width=\linewidth, height=2.8cm,
            keepaspectratio]{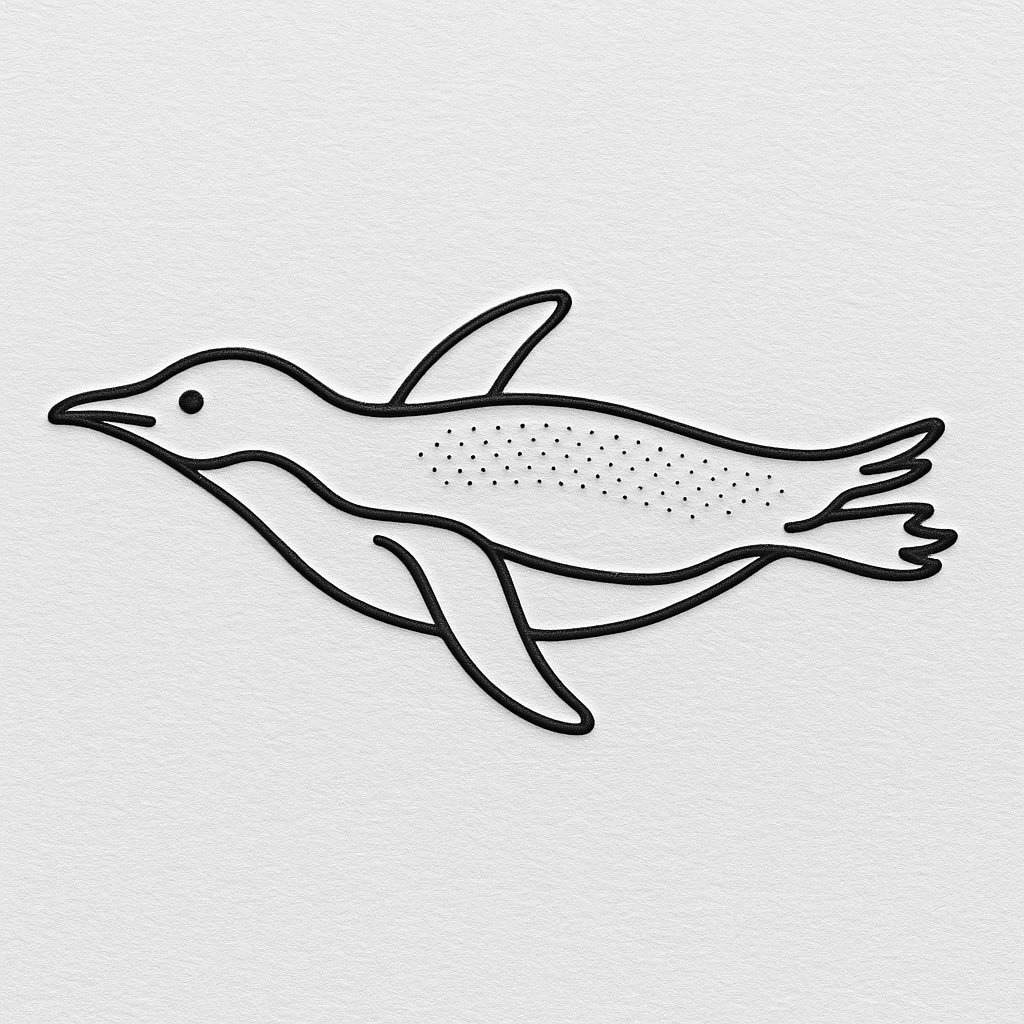}\\[3pt]
        \small (a) Text-only (Penguin)
    \end{minipage}\hfill
    \begin{minipage}[t]{0.48\columnwidth}
        \centering
        \includegraphics[width=\linewidth, height=2.8cm,
            keepaspectratio]{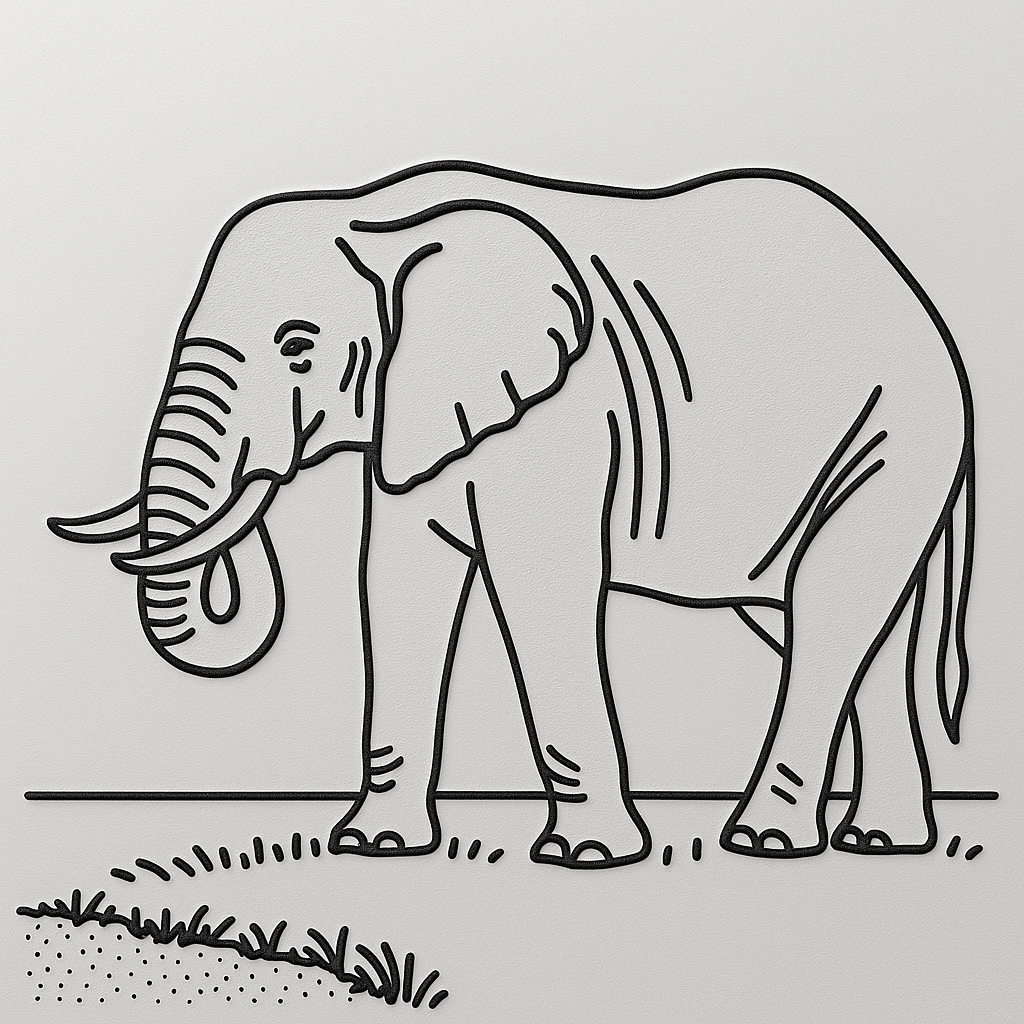}\\[3pt]
        \small (b) Natural-conditioned (Elephant)
    \end{minipage}
    \caption{Zero-shot outputs fail without structured issue guidance: (a)~species
    fidelity breaks; (b)~background clutter persists despite explicit
    removal instructions}
    \label{fig:zeroshot}
\end{figure}

To motivate the structured ViT-driven approach, we first evaluated two
unguided generation modes.
\textbf{Text-only} conditioning produced unreliable species identity: the
Penguin output (Fig.~\ref{fig:zeroshot}a) resembles a marine mammal despite
explicit specification.
\textbf{Natural-conditioned} generation improved species fidelity but
consistently reintroduced background habitat context (Fig.~\ref{fig:zeroshot}b).
Both modes required expert triage before embosser use, confirming that
unstructured generation alone is insufficient.

\subsection{Editing Case Study: F1QL Dinosaur}
\label{sec:casestudy}

\begin{figure}[t]
    \centering
    \begin{minipage}[t]{0.32\columnwidth}
        \centering
        \includegraphics[width=\linewidth, height=3.2cm,
            keepaspectratio]{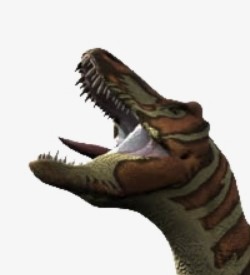}\\[2pt]
        \small (a) Natural Photo
    \end{minipage}\hfill
    \begin{minipage}[t]{0.32\columnwidth}
        \centering
        \includegraphics[width=\linewidth, height=3.2cm,
            keepaspectratio]{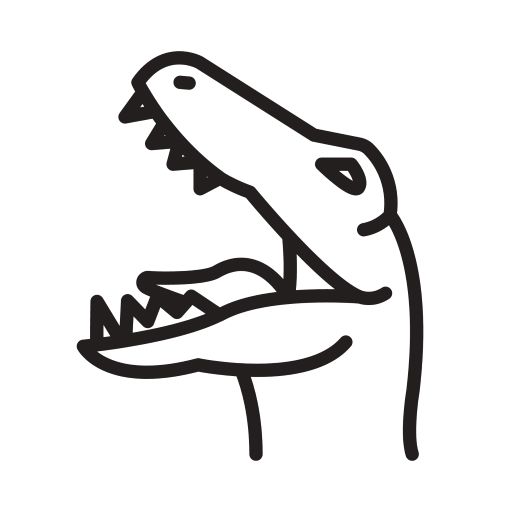}\\[2pt]
        \small (b) Original Tactile
    \end{minipage}\hfill
    \begin{minipage}[t]{0.32\columnwidth}
        \centering
        \includegraphics[width=\linewidth, height=3.2cm,
            keepaspectratio]{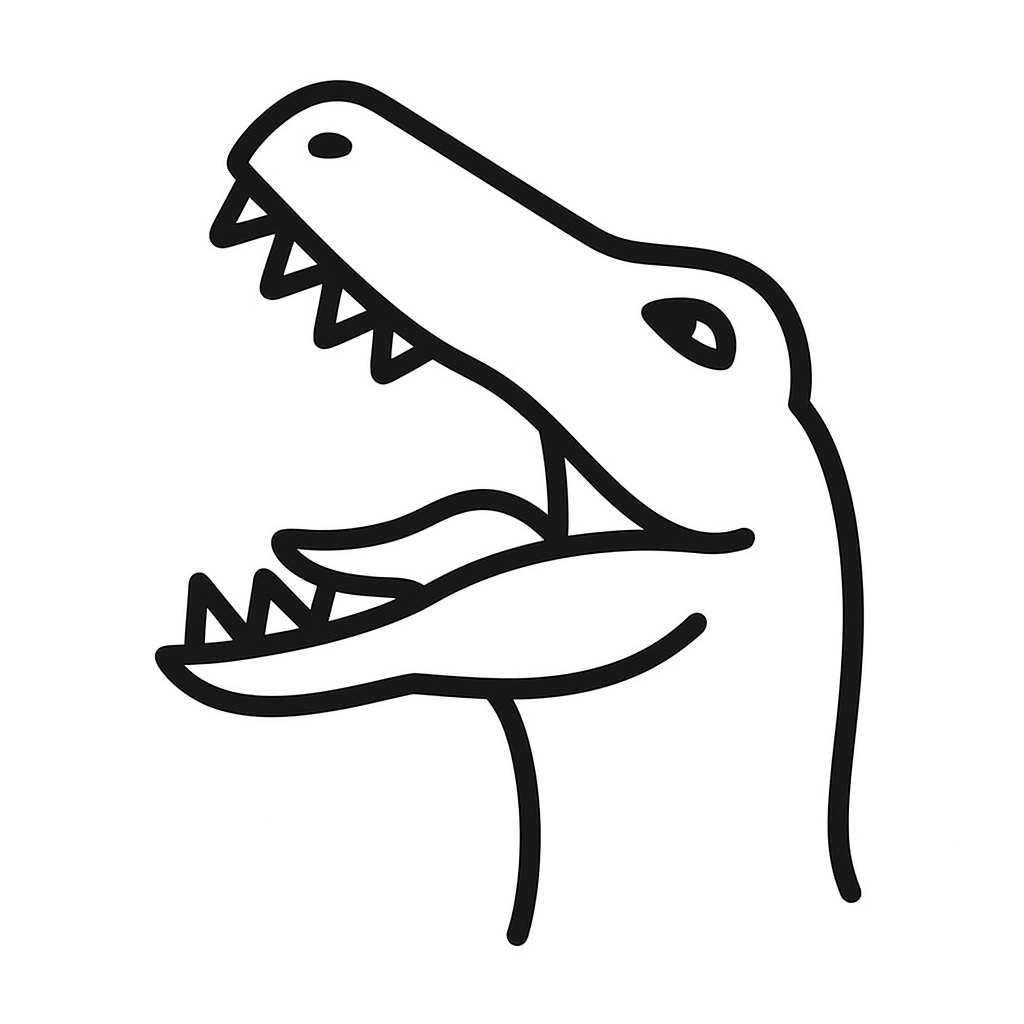}\\[2pt]
        \small (c) ViT Edit
    \end{minipage}
    \caption{F1QL editing case study: Dinosaur.
    (a)~Natural photo reference.
    (b)~Original tactile: excessively thick strokes in teeth.
    (c)~ViT-guided edit (ViT issue prob.\ 0.93 (\texttt{too\_thick}).}
    \label{fig:editing_example}
\end{figure}

Fig.~\ref{fig:editing_example} illustrates the pipeline on a Dinosaur pair from
Family~1 (Animals \& Creatures), Line Quality (F1QL). AMT workers were unanimous
(7/7) in flagging \texttt{too\_thick}; the ViT probe likewise assigned a 93\%
issue probability, so the prompt emphasized thinning oversized strokes and
separating the teeth into clean, distinct contours. The ViT-guided edit
(Fig.~\ref{fig:editing_example}c) clearly reduces the black fills in the mouth
region and restores the outline legibility. Interestingly, the post-edit ViT
score decreases only marginally ($\Delta p = -0.004$, see \ref{edit_case}), highlighting that small
perturbations near the decision boundary do not necessarily match human
perception; the qualitative improvement is obvious even if the classifier’s
confidence barely changes. This underscores why we pair quantitative proxies
with visual inspection in the editing study.

\begin{figure}[t]
    \centering
    \begin{minipage}[t]{0.32\columnwidth}
        \centering
        \includegraphics[width=\linewidth, height=3.2cm, keepaspectratio]{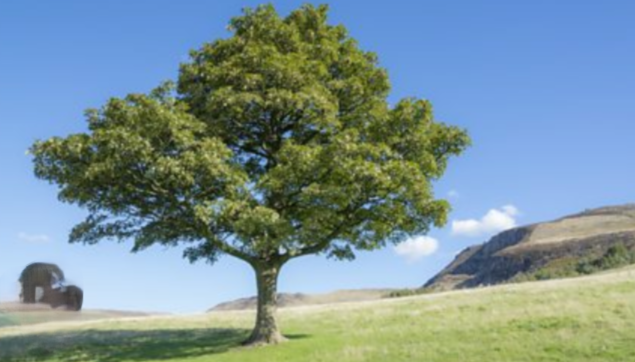}\\[2pt]
        \small (a) Natural Photo
    \end{minipage}\hfill
    \begin{minipage}[t]{0.32\columnwidth}
        \centering
        \includegraphics[width=\linewidth, height=3.2cm, keepaspectratio]{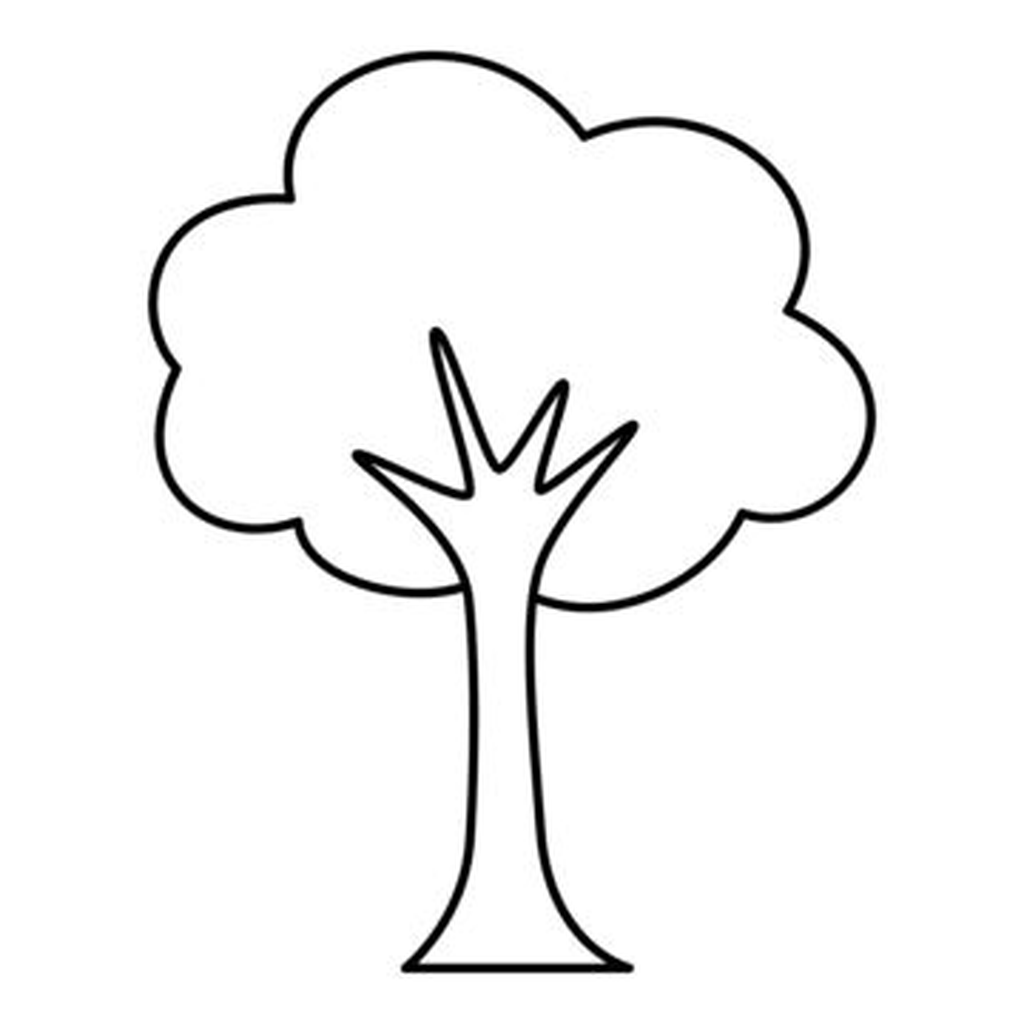}\\[2pt]
        \small (b) Original Tactile
    \end{minipage}\hfill
    \begin{minipage}[t]{0.32\columnwidth}
        \centering
        \includegraphics[width=\linewidth, height=3.2cm, keepaspectratio]{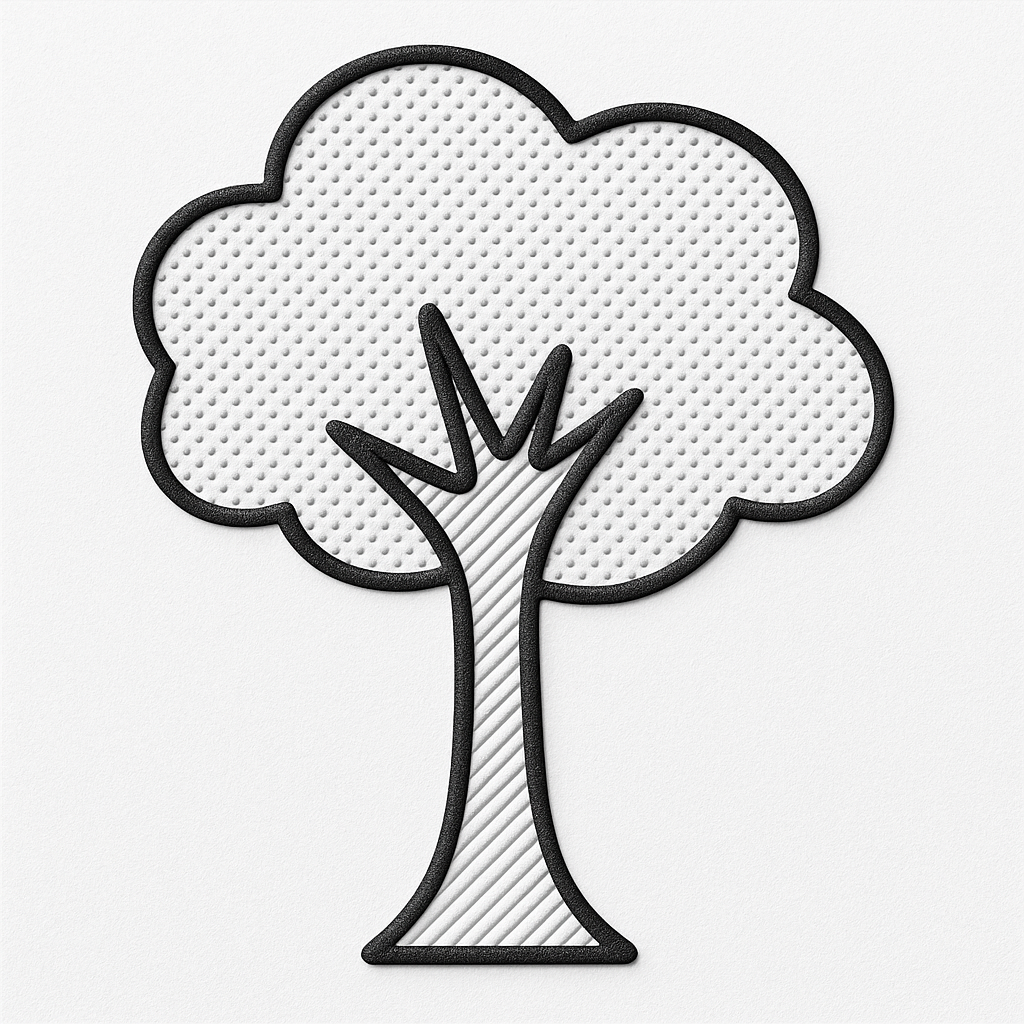}\\[2pt]
        \small (c) ViT Edit
    \end{minipage}
    \caption{F2QT editing case study: Tree (missing texture).
    (a)~Natural photo reference.
    (b)~Original tactile: uniform texture fails to separate canopy regions.
    (c)~ViT-guided edit (issue prob.\ 0.82$\rightarrow$0.33) introduces differentiated fills and clear negative space.}
    \label{fig:editing_tree}
\end{figure}

The Tree case (F2QT) showcases the pipeline on a texture-separation defect. AMT workers (7/7) labeled the tree as \texttt{missing\_texture}, and the ViT probe assigned an 0.82 issue probability. After editing, the model introduces high-contrast fills for the canopy and trunk, carving out negative space that makes each region tactilely distinct. The ViT score drops to 0.33 ($\Delta p =0.485$), aligning with the visual improvement and demonstrating that texture-focused prompts can yield precise, localized edits.

\subsubsection{Quantitative proxy.}
\label{edit_case}
\begin{figure}[t]
    \centering
    \includegraphics[width=\columnwidth]{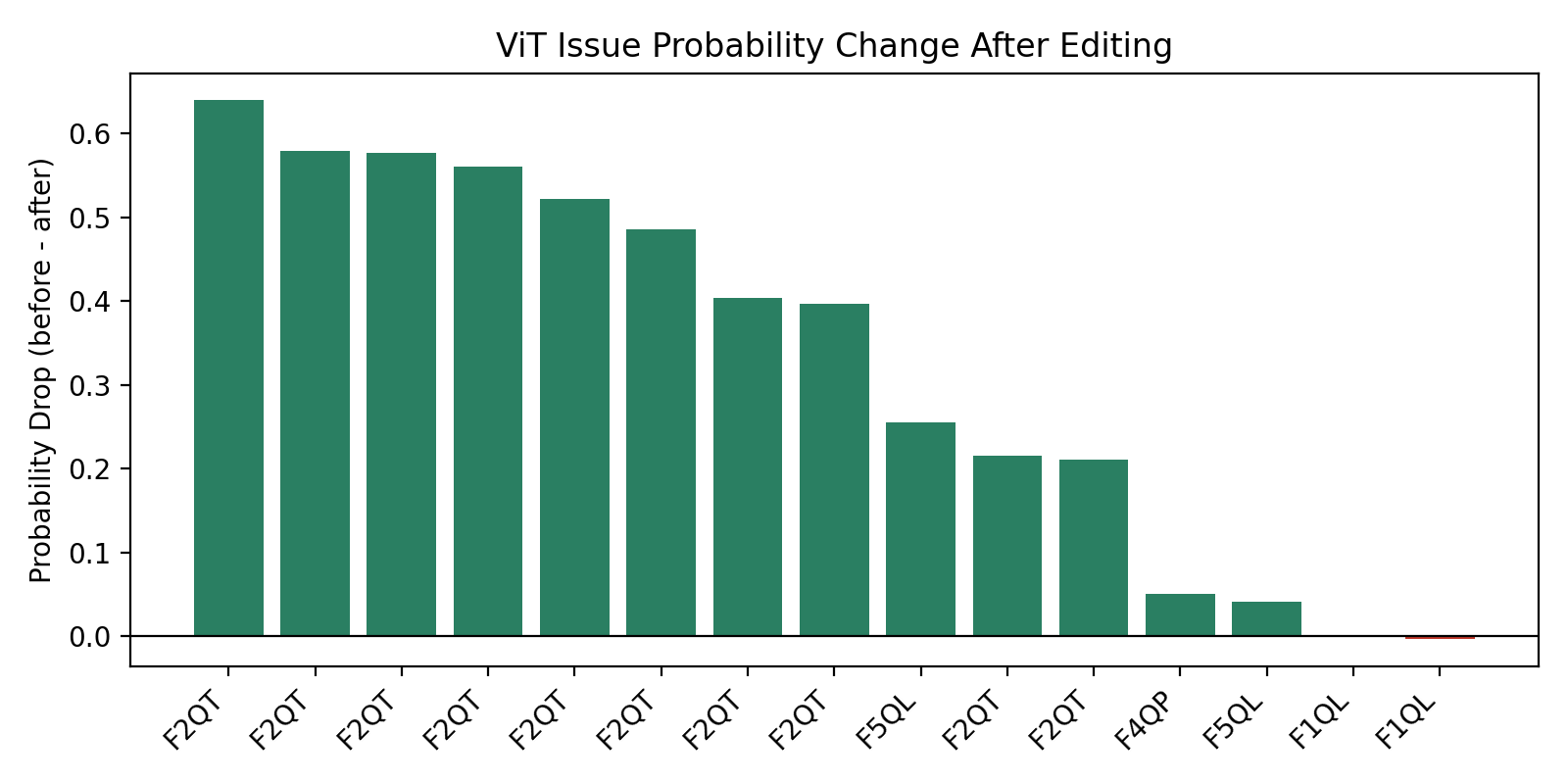}
    \caption{Drop in ViT issue probability (before minus after) for the
    15 high-confidence edits. Positive values indicate improvement; only one
    sample (F1QL dinosaur) slightly worsened.}
    \label{fig:edit_deltas}
\end{figure}

To move beyond single-case anecdotes, we selected the 15 highest-confidence
crowd-supported defects in the test split (all with $\geq5$ worker votes and
ViT issue probability $\geq 0.80$)
and ran the editing pipeline on each.  We then re-scored the edited tactiles
with the \emph{same} ViT probe (no further fine-tuning) to measure the change in
issue probability.  Figure~\ref{fig:edit_deltas} shows the per-sample deltas:
14/15 edits reduced the ViT signal (mean drop 0.329, median 0.397), indicating
that the generated fixes move the render closer to the ``no defect'' manifold.
Table~\ref{tab:edit_proxy} lists representative pairs; note that the only
regression occurred on the F1QL dinosaur example, matching the visual failure
mode discussed earlier.

\begin{table}[t]
\centering
\caption{Before/after ViT probabilities on selected evaluation pairs.}
\label{tab:edit_proxy}
\renewcommand{\arraystretch}{1.1}
\begin{tabular}{lcc@{\,}c}
\toprule
\textbf{Task / Option} & $p_{\text{before}}$ & $p_{\text{after}}$ & $\Delta$ \\
\midrule
F2QT Egg -- \texttt{missing\_texture}        & 0.903 & 0.693 & +0.211 \\
F2QT Planet -- \texttt{missing\_texture} & 0.739 & 0.099 & +0.640 \\
F2QT Tree -- \texttt{missing\_texture}   & 0.699 & 0.121 & +0.577 \\
F5QL Scooty -- \texttt{too\_thick}           & 0.934 & 0.679 & +0.255 \\
F4QP Laptop -- \texttt{missing\_parts}       & 0.858 & 0.807 & +0.050 \\
F1QL Dinosaur -- \texttt{too\_thick}         & 0.985 & 0.989 & $-0.004$ \\
\bottomrule
\end{tabular}
\end{table}

\subsection{Cross-Family Editing Patterns}
\label{sec:cross_family}

Across further experiments spanning all five quality dimensions, consistent
patterns emerge.
In F1QP (Required Parts), high-confidence ViT detections of \texttt{missing\_parts}
yield stable anatomical completions; both limb sets and head restored, while
lower-confidence detections sometimes under-correct.
In F1QT (Texture Separation), the pipeline's weakest dimension in evaluation
(F1QT \texttt{missing\_texture}: 0.68), the editing model tends to over-texture
the full canvas rather than targeting specific body regions, consistent with the
ViT's known difficulty in localizing fill-density defects.
In F1QB (Extra Content) and F1QV (View Match), where ViT accuracy is higher,
the family-level guardrails in the prompt footer provide implicit normalization
beyond the specific repair instruction; background cleanliness and posture
alignment improve even when not the primary target.
These patterns suggest that ViT-guided editing is most reliable when classifier
confidence is high and the defect is spatially localized, and that the
per-family prompt templates provide a useful floor of correction quality even
under imperfect diagnosis.

\section{Discussion}
\label{sec:discussion}

\subsection{Structured Taxonomy as a Scalability Lever}

A key design choice in \textsc{TactileEval} is decomposing expert tactile
knowledge into plain-language, option-level judgments.
This allows non-specialist crowd workers to contribute reliably at scale,
removing the bottleneck of requiring a tactile accessibility expert for every
new annotation cycle.
The 85.70\% overall ViT accuracy across 30 task families confirms that this
structured signal is learnable from crowd labels alone, and the consistent
difficulty ordering (geometric checks easiest, fine-grained texture and part
checks hardest) reflects genuine perceptual structure rather than annotation
noise.

\subsection{ViT Probe as Evaluator and Editing Signal}

The probe's role as an \emph{editing signal} is distinct from its role as a
classifier.
Even when the ViT-selected issue code diverges from the dominant crowd judgment,
the resulting edit still improves the tactile, because the family-level
formatting guardrails in the prompt template apply a consistent floor of
correction quality.
This suggests that issue attribution granularity matters less than prompt quality
for the editing step.

The probe does exhibit systematic probability inflation on high-frequency
training options; notably \texttt{missing\_texture} in F1QT and
\texttt{missing\_parts} across families---leading the ViT channel to
over-generalize toward default repairs rather than instance-specific ones.
Per-option confidence threshold calibration using held-out validation data is
the natural mitigation.

\subsection{Limitations}

First, the \texttt{gpt-image-1} API dependency limits full reproducibility of
the editing step; the ViT probe, dataset, and AMT protocol are fully open.
Second, the editing pipeline currently addresses one diagnosed issue per
invocation and has not been evaluated by BVI participants or domain experts.
The qualitative results in Section~\ref{sec:casestudy} are encouraging but
insufficient to claim embosser-readiness.

\subsection{Future Work}
Per-option confidence threshold calibration using held-out validation splits
per family is a natural next step to address probability inflation on
high-frequency options.
Extending the ViT probe with a lightweight decoder to produce structured
rationales would enable self-contained diagnosis.
A longitudinal study with BVI participants remains essential to validate
whether pipeline-corrected graphics yield measurably better tactile
comprehension outcomes.

\section{Conclusion}
\label{sec:conclusion}

We presented \textsc{TactileEval}, a three-stage pipeline for fine-grained
evaluation and automated editing of tactile graphics.
A five-category BANA-aligned taxonomy operationalizes expert knowledge as
plain-language crowd tasks, enabling a 14,095-record dataset spanning all 66
TactileNet object classes consolidated into six families without requiring specialist annotators at scale.
A ViT-L/14 feature probe trained on this data achieves 85.70\% overall test
accuracy across 30 task families, with difficulty ordering consistent across
families.
A ViT-guided editing pipeline translates probe outputs into targeted
\texttt{gpt-image-1} edits, producing improved tactile renders in
high-confidence, spatially localized defect cases.

We release all data, code, and model configurations and hope this work provides
a concrete foundation for future BVI-validated, fully automated tactile
correction pipelines.

\section*{Acknowledgment}
This work was supported in part by MITACS and the Digital Alliance of Canada.
The authors thank the student volunteers at the Intelligent Machines Lab (iML),
Carleton University, for their contributions, Joshua Olojede and Hoda Vafaeesefat for their help
with the AMT annotation environment.

\bibliographystyle{ieeetr}
\bibliography{references}

\end{document}